\documentclass[letterpaper, 10 pt, conference]{ieeeconf}  

\IEEEoverridecommandlockouts                              

\usepackage{epsfig} 
\usepackage{graphicx}
\usepackage{amsmath} 
\usepackage{amssymb}  
\usepackage{booktabs}
\usepackage{xcolor}

\newcommand{\ie}[0]{\textit{i.e.} }
\newcommand{\eg}[0]{\textit{e.g.} }

\title{\LARGE \bf
Temporal Context for Robust Maritime Obstacle Detection 
}

\author{Lojze Žust$^{1}$ and Matej Kristan$^{1}$
\thanks{*This work was supported by the Slovenian research agency program P2-0214 and project J2-2506.}
\thanks{$^{1}$Lojze Žust and Matej Kristan are with University of Ljubljana, Faculty of Computer and Information Science, Slovenia
        {\tt\small \{lojze.zust, matej.kristan\}@fri.uni-lj.si}}%
}

\usepackage{tikz}
\usepackage{textcomp}
\usepackage{lipsum}

\newcommand\copyrighttext{%
  \footnotesize \textcopyright 2022 IEEE. Personal use of this material is permitted. Permission from IEEE must be obtained for all other uses, in any current or future media, including reprinting/republishing this material for advertising or promotional purposes, creating new collective works, for resale or redistribution to servers or lists, or reuse of any copyrighted component of this work in other works.}
\newcommand\copyrightnotice{%
\begin{tikzpicture}[remember picture,overlay]
\node[anchor=south,yshift=10pt] at (current page.south) {\fbox{\parbox{\dimexpr\textwidth-\fboxsep-\fboxrule\relax}{\copyrighttext}}};
\end{tikzpicture}%
}

\makeatletter
\let\NAT@parse\undefined
\makeatother
\usepackage[colorlinks=true]{hyperref}

\begin{document}

\maketitle
\copyrightnotice
\thispagestyle{empty}
\pagestyle{empty}

\begin{abstract}
Robust maritime obstacle detection is essential for fully autonomous unmanned surface vehicles (USVs). 
The currently widely adopted segmentation-based obstacle detection methods are prone to misclassification of object reflections and sun glitter as obstacles, producing many false positive detections, effectively rendering the methods impractical for USV navigation.
However, water-turbulence-induced temporal appearance changes on object reflections are very distinctive from the appearance dynamics of true objects. We harness this property to design WaSR-T, a novel maritime obstacle detection network, that extracts the temporal context from a sequence of recent frames to reduce ambiguity. 
By learning the local temporal characteristics of object reflection on the water surface, WaSR-T substantially improves obstacle detection accuracy in the presence of reflections and glitter.
Compared with existing single-frame methods, WaSR-T reduces the number of false positive detections by 41\% overall and by over 53\% within the danger zone of the boat, while preserving a high recall, and achieving new state-of-the-art performance on the challenging MODS maritime obstacle detection benchmark. The code, pretrained models and extended datasets are available at: \url{https://github.com/lojzezust/WaSR-T}
\end{abstract}

\section{Introduction}

Advances in maritime robotics over the last two decades have fostered an emergence of unmanned surface vehicles (USVs). These autonomous boats range from small vessels used for automated inspection of dangerous areas and automation of repetitive tasks like bathymetry or environmental control, to massive cargo and transport ships. This next stage of maritime automation holds a potential to transform maritime-related tasks and will likely impact the global economy. The safety of autonomous navigation systems hinges on their environment perception capability, in particular obstacle detection, which is responsible for timely reaction and collision avoidance.


Cameras as low-power and information rich sensors are particularly appealing due to their large success in perception for autonomous cars~\cite{Cordts2016Cityscapesa,Chen2018Encoder}. However, recent works~\cite{Bovcon2019Mastr,Bovcon2020MODS} have shown that methods developed for autonomous cars do not translate well to USVs due to the specifics of the maritime domain. As a result, several approaches that exploit the domain specifics for improved detection accuracy have been recently proposed~\cite{Bovcon2021WaSR,Cane2019Evaluating,Yao2021Shoreline,Zust2021SLR}. Since everything but water can be an obstacle, classical detectors for individual obstacle classes cannot address all obstacle types. State-of-the-art methods~\cite{Bovcon2021WaSR} instead casts maritime obstacle detection as an anomaly segmentation problem by segmenting the image into the water, sky and obstacle categories.
\begin{figure}[t]
\centering
\includegraphics[width=\linewidth]{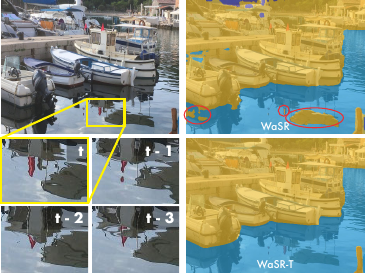}
\caption{Single-frame obstacle detection methods (top right) struggle to distinguish between object reflections and true objects. However, reflections exhibit a distinctive temporal pattern compared to true objects (bottom left). WaSR-T (bottom right) considers the temporal context from recent frames to learn these patterns and increase segmentation robustness.}
\label{fig:motivation}
\end{figure}

Despite significant advances reported in the recent maritime benchmark~\cite{Bovcon2020MODS}, the state-of-the-art is still challenged by the reflective properties of the water surface, which cause objects reflections and sun glitter. In fact, given a single image, it is quite difficult to distinguish a reflected object or a spot of sun glitter from a true obstacle (Figure~\ref{fig:motivation}). This results in a number of false positive detections, which in practice leads to frequent and unnecessary slowdowns of the boat, rendering current camera-based obstacle detection methods impractical.


We note that while correctly classifying reflections from a single image is challenging, the problem might become simpler when considering the temporal context. As illustrated in Figure~\ref{fig:motivation}, due to water dynamics, the reflection appearance is not locally static, like that of an obstacle, but undergoes warped deformations. Based on this insight, we propose a new maritime obstacle segmentation network WaSR-T, which is our main contribution. WaSR-T introduces a new temporal context module that allows the network to extract the temporal context from a sequence of frames to differentiate objects from reflections. To the best of our knowledge, this is the first deep maritime obstacle detection architecture with a temporal component.

\begin{figure*}
\centering
\includegraphics[width=\linewidth]{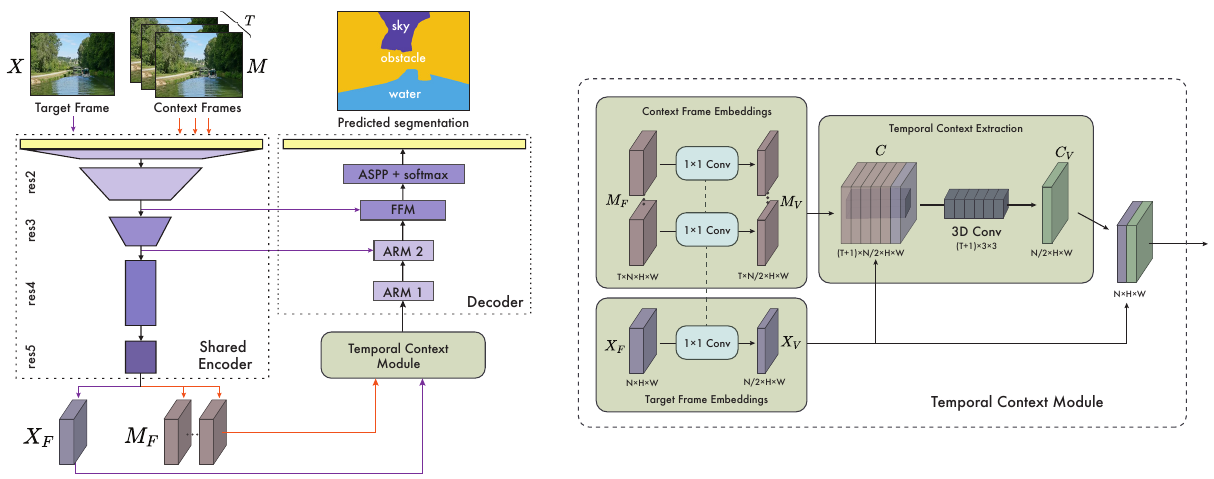}
\caption{Overview of WaSR-T (left). Target frame and preceding context frames are fed into a shared encoder producing per-frame feature maps $X_F$ and $M_F$. The Temporal Context Module (right) extracts the temporal information from per-frame embeddings using a 3D convolution. The resulting temporal context embeddings $C_V$ are combined with target frame embeddings $X_V$ and fed into the decoder which predicts the target frame segmentation.}
\label{fig:architecture}
\end{figure*}


We also observe that the challenging maritime mirroring and glitter scenes are underrepresented in the standard training sets. We therefore extend the existing single-frame maritime segmentation training dataset MaSTr1325~\cite{Bovcon2019Mastr} with corresponding preceding frames and introduce additional training images representing challenging reflection conditions, which is our secondary contribution. To maintain the notation convention, we name the extended dataset MaSTr1478. Experiments show that the dataset extension delivers significant performance improvement. Results on the recent maritime benchmark MODS~\cite{Bovcon2020MODS} show that, compared to the single-frame WaSR~\cite{Bovcon2021WaSR}, the proposed \mbox{WaSR-T} reduces the number of false positive detections by 30\% with a low computational overhead and sets a new state-of-the-art in maritime obstacle detection.

In summary, our main contributions are: (i) WaSR-T, a temporal extension of WaSR~\cite{Bovcon2021WaSR} that leverages the temporal context for increased robustness and (ii) MaSTr1478, an extension of the existing single-frame training dataset~\cite{Bovcon2019Mastr} with challenging reflection scenes that facilitates the training of temporal maritime segmentation networks. The new dataset and the WaSR-T source code will be publicly released to facilitate further research of temporal features in maritime obstacle detection.

\section{Related work}


Semantic segmentation has become a common approach for obstacle detection in the marine domain~\cite{Bovcon2019Mastr,Cane2019Evaluating,Bovcon2020MODS}, as it can address both dynamic (\eg boats, swimmers, buoys) and static obstacles (\eg shore, piers, floating fences) in a unified way by posing the problem as anomaly segmentation. Recently, several specialized networks for the marine domain have been proposed for this task~\cite{Bovcon2021WaSR,Yao2021Shoreline,Qiao2022Automated}. These methods address reflections and increase detection robustness in multiple ways, including regularization techniques~\cite{Yao2021Shoreline}, specialized loss functions~\cite{Bovcon2021WaSR} and obstacle-oriented training regimes~\cite{Zust2021SLR}. 

However, robustness to reflections is still lacking and causes comparatively low performance within the 15m area near the boat~\cite{Bovcon2020MODS}, where segmentation errors are most critical. In practice, obstacle detection methods receive frames sequentially, thus the temporal component of the data is also available and could be used to distinguish between reflections and objects. So far, the additional temporal information has not yet been explored in context of maritime obstacle detection.

In other domains with similar access to sequential image data, effort has been made to harness the temporal information to improve the segmentation performance. Some approaches investigate the use of temporal information only during training to improve the temporal consistency of single-frame networks. \cite{Varghese2021Unsupervised} and \cite{Liu2020Efficient} achieve this by propagating the segmentation masks in consecutive frames by optical flow.

Incorporating temporal information into the network for improved prediction has been explored as well,
with attention-based approaches being the most prevalent method. In video object segmentation~\cite{Oh2019Video,Li2020Fast,Duke2021SSTVOS} attention is used to aggregate the information from features and segmentation masks of previous "memory" frames based on the attention between the target and memory features. However, these methods are designed mainly for propagating initial segmentation masks of large foreground objects over the video sequence and are not directly suitable for general purpose discriminative semantic segmentation required for obstacle detection.

Similarly, in video semantic segmentation~\cite{Wang2021Temporal,Yuan2021CSANet} attention-based approaches are used to aggregate the temporal information from recent frames to improve general purpose semantic segmentation. \cite{Yuan2021CSANet} additionally introduces auxiliary losses, which guide the learning of attention based on inter-frame consistency. Instead of a global attention which aggregates information from semantically similar regions from past frames, we propose a convolutional approach to facilitate the learning of local temporal texture, which is characteristic for reflections.

\section{Temporal context for obstacle detection}


Given a target frame $X \in \mathbb{R}^{3 \times H \times W}$, the task of the segmentation-based obstacle detection method is to predict the segmentation mask, \ie to classify each location in $X$ as either water, sky or obstacle. We propose using the temporal context to improve the prediction accuracy. Our network (Figure~\ref{fig:architecture}), denoted as \mbox{WaSR-T}, is based on the state-of-the-art single-frame network for maritime obstacle detection
WaSR~\cite{Bovcon2021WaSR}. 
We design WaSR-T to encode the discriminative temporal information about local appearance changes of a region over $T$ preceding context frames $M \in \mathbb{R}^{T \times 3 \times H \times W}$. The temporal context is added to the high-level features at the deepest level of the network as shown in Figure~\ref{fig:architecture}.

Following \cite{Oh2019Video} and \cite{Wang2021Temporal}, the target and context frames are first individually encoded with a shared encoder network, producing per-frame feature maps $X_F \in \mathbb{R}^{N \times H \times W}$ and $M_F \in \mathbb{R}^{T \times N \times H \times W}$, where $N$ is the number of channels.
The Temporal Context Module (Section~\ref{sec:method/temporal_descriptors}) then extracts dicriminative temporal context embeddings from per-frame features. Finally, the temporal context embeddings are concatenated with target frame embeddings and fed into a decoder network. Following \cite{Bovcon2021WaSR}, the decoder gradually merges the information with multi-level features of the target frame (\ie skip connections) and outputs the final target frame segmentation.

\subsection{Temporal Context Module}\label{sec:method/temporal_descriptors}

The Temporal Context Module (TCM) extracts the temporal information from embeddings of the context and target frames and combines it with embeddings of the target frame using concatenation (Figure~\ref{fig:architecture}). For this reason, the number of input channels to the decoder doubles compared to the single-frame network. Thus, in order to preserve the structure and number of input channels to the decoder, TCM first reduces the dimensionality of per-frame feature maps $X_F$ and $M_F$ accordingly -- a shared $1 \times 1$ convolutional layer is used to project the per-frame feature maps into $N/2$ dimensional per-frame embeddings $X_V$ and $M_V$ as shown in Figure~\ref{fig:architecture}.

To extract the temporal information from a sequence of frame embeddings, attention-based approaches~\cite{Oh2019Video,Duke2021SSTVOS,Wang2021Temporal} are often utilized, as they allow aggregation of information from semantically similar regions across multiple frames to account for movement and appearance changes of objects. Reflections, however often feature significant local texture changes as demonstrated in Figure~\ref{fig:motivation}.
Thus, instead of globally aligning semantically similar regions using attention mechanisms, we utilize a spatio-temporal convolution to extract the local texture changes.

First we stack the context and target frame embeddings $M_V$ and $X_V$ into a spatio-temporal context volume $C \in \mathbb{R}^{(T+1) \times N/2 \times H \times W}$. Then a 3D convolution layer is used to extract discriminative spatio-temporal features from $C$. To account for minor inter-frame object and camera movements, a kernel size of $(T+1) \times 3 \times 3$ is used to capture temporal information in a local spatial region around locations in the context volume. We apply padding in the spatial dimensions to preserve the spatial size of the output context features $C_V  \in \mathbb{R}^{N/2 \times H \times W}$.

\subsection{Efficient inference}

During training, for each input image $X$, WaSR-T needs to extracts all per-frame context embeddings $M_F$ in addition to target frame embeddings $X_V$. However, during inference the frames are passed to the network sequentially, thus recent frame embeddings can be stored in memory and feature extraction only needs to be performed on the newly observed target frame. Specifically, WaSR-T stores a buffer of $T$ most recent frame embeddings $X_V$ in memory and uses them as the context frame embeddings $M_V$ in TCM. The memory buffer is initialized with $T$ copies of the $X_V$ embeddings of the first frame in the sequence. Using sequential inference, the efficiency of WaSR-T is not significantly impacted compared to single-frame methods, differing only due to the temporal context extraction in TCM.





\section{Experiments}

\begin{figure*}
\centering
\includegraphics[width=\linewidth]{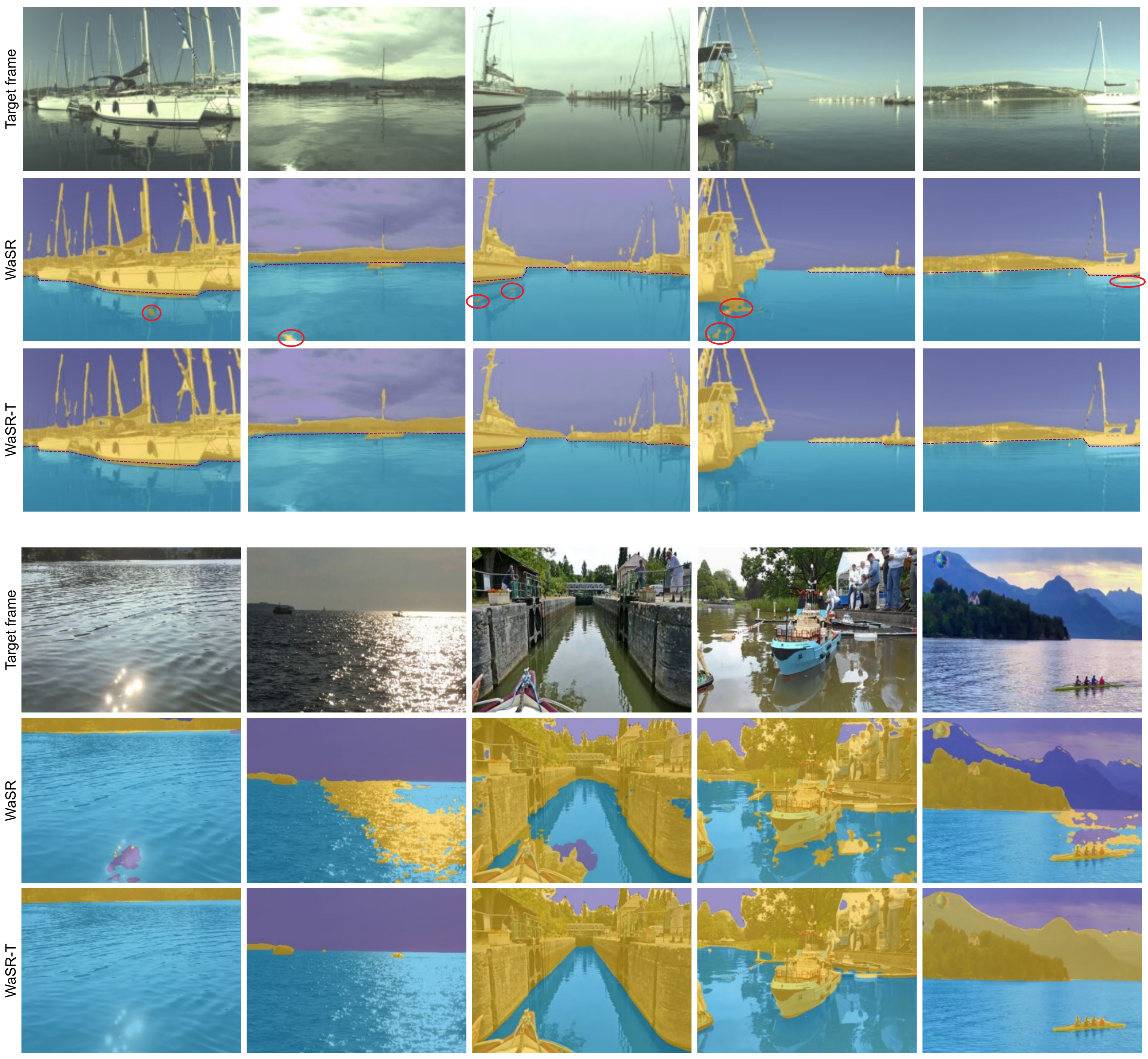}
\caption{Qualitative results on MODS (top) and web-sourced sequences (bottom) reveal that in WaSR-T the handling of reflections and sun glitter is significantly improved compared to WaSR, resulting in a smaller number of FP detections and increased temporal consistency.}
\label{fig:cmp_mods}
\end{figure*}

\subsection{Implementation details}

WaSR-T follows the architecture of WaSR~\cite{Bovcon2021WaSR} and applies ResNet101 as the feature encoder. In a preliminary study we observed that in contrast to WaSR, the inertial measurements (IMU) do not bring improvements in our temporal extension. Therefore the IMU is not used in the decoder for simplicity. We apply the original WaSR training procedure, i.e., the water separation loss function, hyper-parameters, optimizers, learning rate schedule and image augmentation. We set the number of past frames in the temporal context module to $T=5$. Because of training memory constraints, the backbone gradients are restricted to the current and previous frame. WaSR-T is trained for 100 epochs on 2$\times$NVIDIA Titan A100S GPUs with a minibatch size of 4 per GPU.

The networks in our experiments are trained on the training set Mastr1478 (Section~\ref{sec:mastr1478}) and tested on the most recent maritime obstacle detection benchmark MODS~\cite{Bovcon2020MODS}, which contains approximately 100 annotated sequences captured under various conditions. The evaluation protocol reflects the detection performance meaningful for practical USV navigation and separately evaluates the detection of obstacle-water edge for static obstacles and the detection of dynamic obstacles. The water-edge detection robustness ($\mu_R$) is computed from the ground truth edge, while dynamic obstacle detection is evaluated in terms of true-positive (TP), false-positive (FP) and false-negative (FN) detections, and summarized by the F1 measure, precision (Pr) and recall (Re). A dynamic obstacle counts as detected (TP) if the coverage of the segmentation inside the ground truth bounding box is sufficient, otherwise the obstacle counts as undetected (FN). Predicted segmentations outside of the ground truth bounding boxes count as false positive detections. Detection performance is reported over the entire visible navigable area and separately within a 15m \textit{danger zone} from the USV, where the detection performance is critical for immediate collision prevention.

\subsection{Temporally extended training dataset MaSTr1478}\label{sec:mastr1478}

\begin{figure}
\centering
\includegraphics[width=\linewidth]{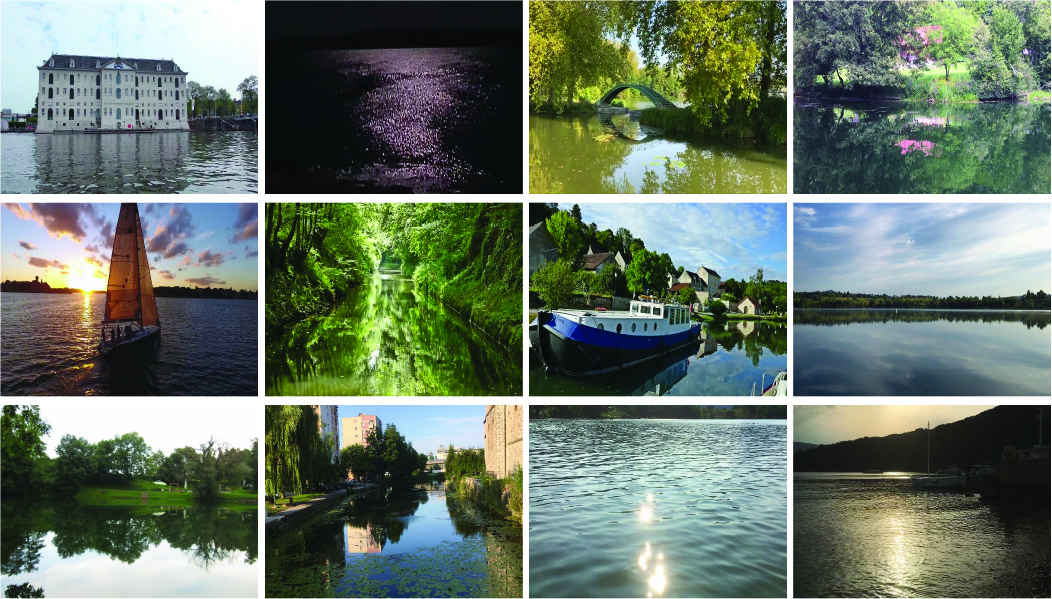}
\caption{Examples of the additional training sequences in the MaSTr1478 with object reflections, sun glitter and low-light conditions.}
\label{fig:dataset}
\end{figure}

To facilitate the training of temporal networks, we extended the recent MaSTr1325~\cite{Bovcon2019Mastr} dataset, which contains 1325 fully segmented images recorded by a USV. First, the dataset was extended by adding $T=5$ preceding frames for each annotated frame, to allow learning of the temporal context. We noticed that while MaSTr1325 is focused on the broader challenges in maritime segmentation, it contains relatively few examples of challenging reflections and glitter. We have thus extended the original dataset with additional 153 images (including their preceding frames) and use the codename \textit{MaSTr1478} for this new dataset. The additional images were obtained from online videos or were additionally recorded by us to represent difficult scenarios for current single-frame methods, where the temporal information is important for accurate prediction, such as object mirroring, reflections and sun glitter. Examples are shown in Figure~\ref{fig:dataset}. The frames are labeled with per-pixel ground truth following~\cite{Bovcon2019Mastr}. To emphasize the challenging conditions, the training samples in the training batches are sampled from the original MaSTr1325 images and the additional images with equal probability. 

\subsection{Comparison with state of the art}

We compare WaSR-T with single-frame state-of-the-art segmentation methods (DeepLabV3+~\cite{Chen2018Encoder}, BiSeNet~\cite{Yu2018Bisenet}, RefineNet~\cite{Lin2017RefineNet}, WaSR~\cite{Bovcon2021WaSR}), which scored as top performers on the recent maritime obstacle detection benchmark MODS~\cite{Bovcon2020MODS}, as well as with state-of-the-art segmentation methods that rely on temporal information. For the latter we considered the video object segmentation method STM~\cite{Oh2019Video} and a recent video semantic segmentation method TMANet~\cite{Wang2021Temporal}, which use memory attention to encode the temporal information from past frames. We use our implementation of STM, which is based on the WaSR network and extends it with the memory attention mechanism of STM~\cite{Bovcon2021WaSR}.



\begin{figure*}
\centering
\includegraphics[width=\linewidth]{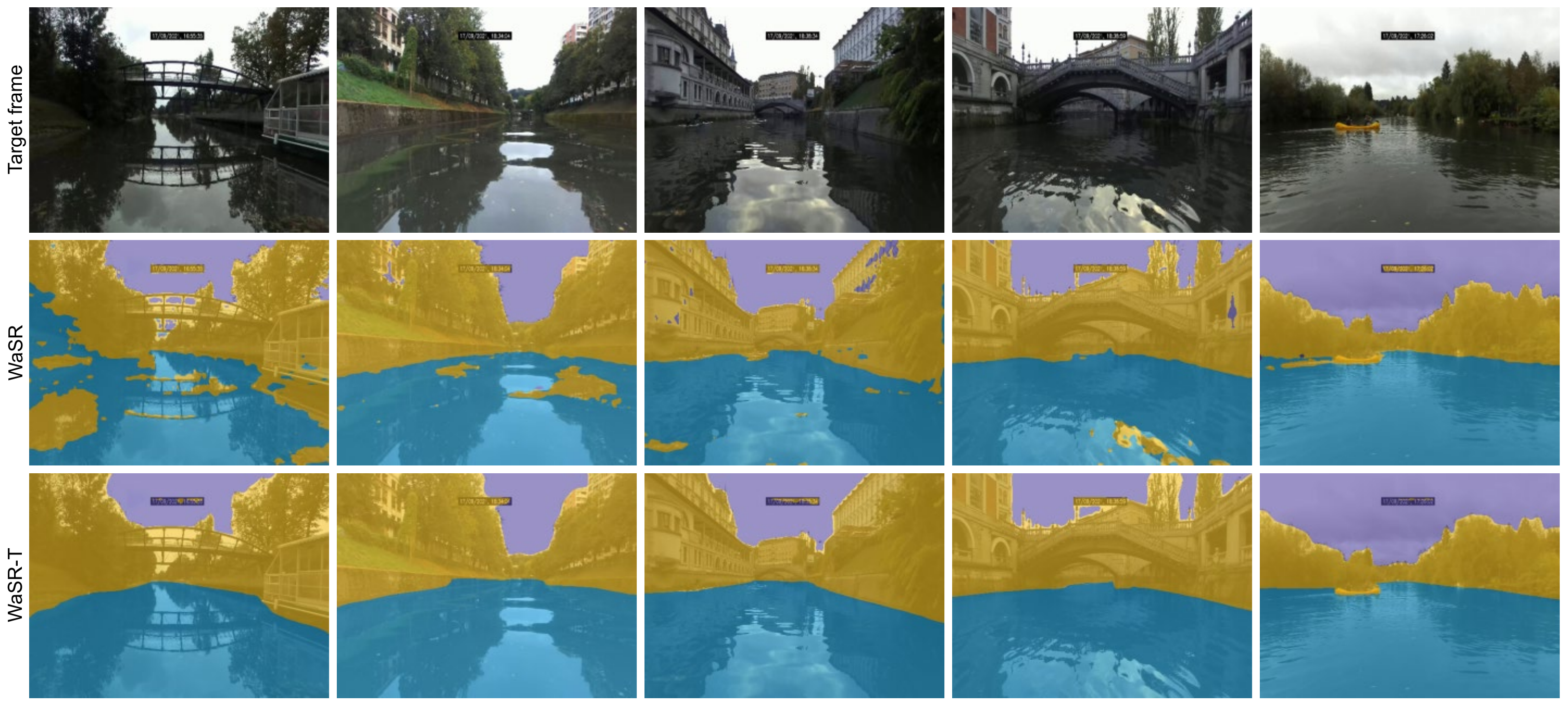}
\caption{Qualitative results on challenging inland water sequences demonstrates large improvements of WaSR-T in terms of practical robustness to reflections.}
\label{fig:cmp_davimar}
\end{figure*}

Results in Table~\ref{tab:sota} show that multi-frame methods outperform the single-frame networks in detection precision (particularly within the danger zone), and except from TMANet, preserve a high recall.
WaSR-T outperforms the original WaSR by 1.8 points in precision and 0.9 points in the overall F1, while substantially outperforming it within the danger zone resulting in a 6.0 points F1 score improvement. This is primarily due to reduction of false positives (see Figures \ref{fig:cmp_mods} and \ref{fig:cmp_davimar}), which is reflected in a 10.5 point improvement of the Pr score within the danger zone. Importantly, the increase in precision does not come at a cost in recall, which remains almost identical to the single-frame counterpart. \mbox{WaSR-T} also outperforms the temporal state-of-the-art networks especially inside the danger zone, resulting in approximately 2 points performance improvement of danger-zone F1 score.

In terms of speed, the new temporal module does not substantially increase the computation. The original WaSR runs at 15 FPS, while WaSR-T runs at approximately 10 FPS, which matches the sequence acquisition framerate.

\begin{figure*}
\centering
\includegraphics[width=\linewidth]{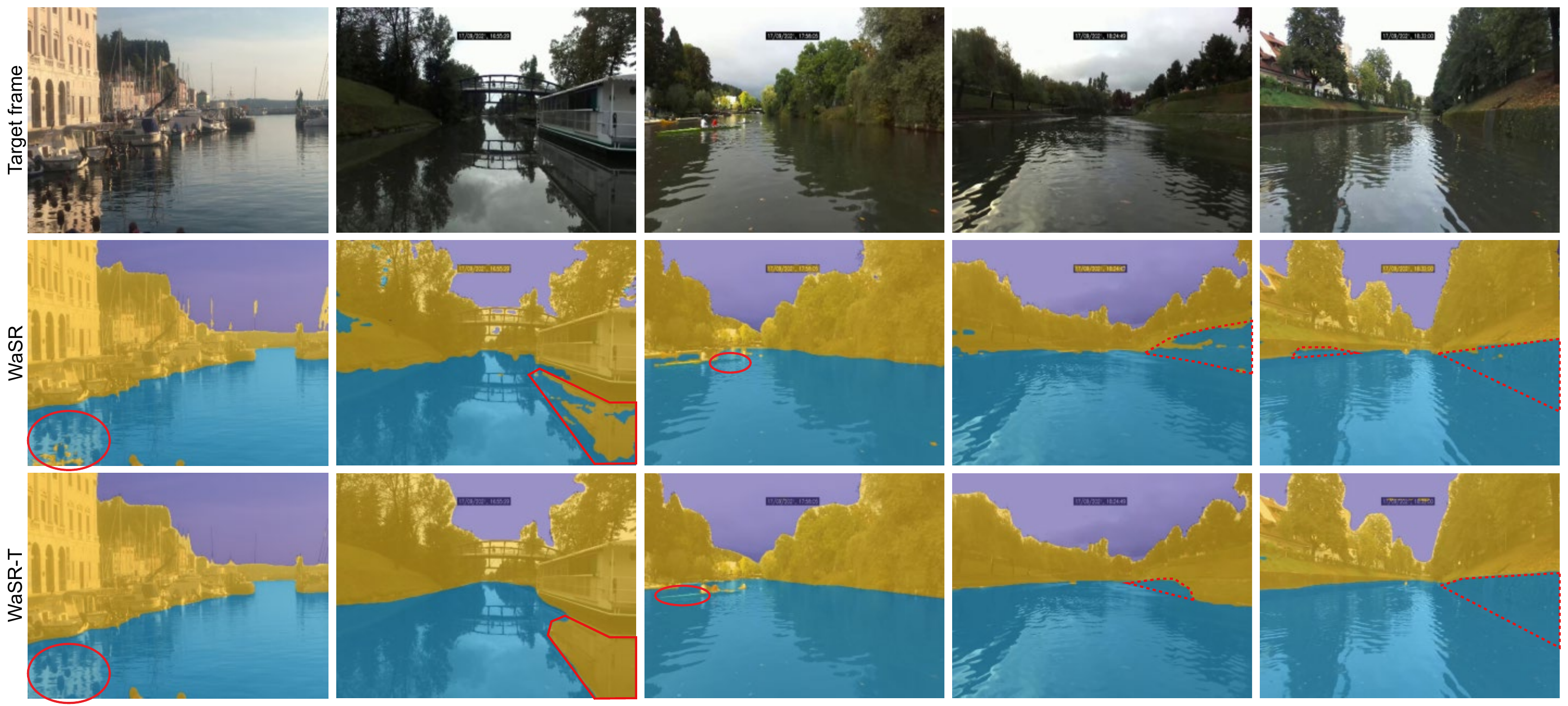}
\caption{Failure cases of both methods include small objects hiding in reflections (column 1), reflections on very still water (column 2), thin objects (column 3) and challenging water-land boundaries (columns 4 and 5).}
\label{fig:cmp_failure}
\end{figure*}

Despite the large improvements in robustness to reflections, WaSR-T also shares some limitations (\eg detection of thin objects) with existing methods as shown in Figure~\ref{fig:cmp_failure}. For example, the temporal context is still not able to fully address reflections in rare situations where the water is completely still and the temporal texture changes cannot be observed. We aim to tackle these challenges in our future work. 

\begin{table}[]
    \centering
    \caption{Comparison of SOTA single-frame and multi-frame methods on MODS in terms of water-edge detection robustness ($\mu_R$), precision, recall and F1 score for obstacle detection. Danger-zone performance is reported in parentheses.}
    \label{tab:sota}
    \begin{tabular}{lcccc}
    method    & $\mu_R$ &   Pr & Re  &      F1 \\
    \midrule
    DeepLabV3+~\cite{Chen2018Encoder}  & 96.8 & 80.1 (18.6) & \textbf{92.7} (\textbf{98.4}) & 86.0 (31.3) \\
    BiSeNet~\cite{Yu2018Bisenet}     & 97.4 & 90.5 (53.7) & 89.9 (97.0) & 90.2 (69.1) \\
    RefineNet~\cite{Lin2017RefineNet}   & 97.3 & 89.0 (45.1) & 93.0 (98.1) & 91.0 (61.8) \\
    WaSR~\cite{Bovcon2021WaSR}   &  97.8 & 95.1 (80.3) &  91.9 (96.2) &  93.5 (87.6) \\
    \midrule
    TMANet~\cite{Wang2021Temporal} &  98.3 &  96.4 (90.0) &  85.1 (93.0) &  90.4 (91.5) \\
    STM~\cite{Oh2019Video}    &  \textbf{98.4} &  96.3 (86.2) &  92.5 (96.4) &  \textbf{94.4} (91.0) \\
    WaSR-T &  \textbf{98.4} &  \textbf{96.9} (\textbf{90.8}) &  92.0 (96.5) &  \textbf{94.4} (\textbf{93.6}) \\
    \end{tabular}
\end{table}

\subsection{Analysis of the alternative temporal aggregation methods}

Next, we analyzed alternatives to the feature fusion in the temporal context module proposed in Section~\ref{sec:method/temporal_descriptors}: (i) pixel-wise average pooling of temporal features (window size of $T + 1 \times 1 \times 1$) and (ii) local average pooling of temporal features ($T + 1 \times 3 \times 3$). Table~\ref{tab:temp_agg} shows that, compared to singe-frame WaSR, the simple pixel-wise temporal average pooling of context features already improves the performance over single-frame inference by 0.8 points (overall) and 1.9 points (danger zone) in F1. Increasing the pooling window size to a local window does not improve performance. In contrast, the 3D convolution approach described in Section~\ref{sec:method/temporal_descriptors} is able to learn discriminative local temporal relations and increases the F1 by an additional 0.2 points overall, and by 3.5 points inside the danger zone. The improvement is primarily on the account of substantial reduction of false positive detection while retaining a high number of true positive detections.
\begin{table}[htb]
    \centering
    \caption{WaSR-T performance with different temporal aggregation methods in terms of water-edge detection robustness ($\mu_R$), number of TP and FP detections and F1 score. Performance inside the danger-zone is reported in parentheses.}
    \label{tab:temp_agg}
    \begin{tabular}{lcccc}
    aggregation        & $\mu_R$ &   TP &      FP &       F1 \\
    \midrule
    Single-frame       &  97.8 & 48241 (2567) & 2492 (629) &  93.5 (87.6) \\
    Avg pool$_{1 \times 1}$    &  \textbf{98.4} & \textbf{48314} (2575)  & 1771 (474) &  94.2 (90.1) \\
    Avg pool$_{3 \times 3}$    &  98.3 & 48011 (\textbf{2578}) &  2152 (537) &  93.5 (89.2) \\
    3D conv    &  \textbf{98.4} & 48284 (2575) & \textbf{1540} (\textbf{261}) &  \textbf{94.4} (\textbf{93.6}) \\
    \end{tabular}
\end{table}

\subsection{Influence of the temporal and spatial context size}

To gain further insights, we analyzed the influence of temporal context module parameters, i.e., the temporal context length $T$ and spatial kernel size. Table~\ref{tab:abl} shows that utilizing even a single temporal context frame (i.e., $T=1$) significantly improves the performance over single-frame inference ($T=0$) by decreasing the number of false positive detections by 30\% overall and 39\% inside the danger zone. This shows the importance of the temporal context in this domain and is somewhat expected as usually a single frame is enough to observe the temporal texture changes common in reflections (see Figure~\ref{fig:motivation}).
Increasing the temporal context length $T$ further, brings consistent, but smaller improvements in reduction of FP detections and danger-zone F1 scores.


The spatial context size, determined by the kernel size of the 3D convolution of the temporal context module also importantly affects the performance. Using $1 \times 1$ spatial kernel size encodes only pixel-wise temporal relations, which negatively impacts the performance inside the danger-zone within which the objects are typically large. Increasing the kernel size to $3 \times 3$ addresses this issue, while the performance does not improve with further increasing the spatial context size. We hypothesize this is due to the local nature of texture changes that occur on reflections, rendering larger kernel sizes redundant.
\begin{table}[htb]
    \centering
    \caption{Influence of parameters in WaSR-T in terms of water-edge detection robustness ($\mu_R$), number of FP detections and F1 score. Performance inside the danger-zone is reported in parentheses.}
    \label{tab:abl}
    \begin{tabular}{cccc}
    $T$    & $\mu_R$ &         FP &       F1 \\
    \midrule
    0      &  97.8 &  2492 (629) &  93.5 (87.6) \\
    1      &  98.4 &  1745 (383) &  94.2 (91.5) \\
    3      &  \textbf{98.6} &  1606 (323) &  94.0 (92.6) \\
    5      &  98.4 &  \textbf{1540} (\textbf{261}) &  \textbf{94.4} (\textbf{93.6}) \\
    \midrule
    kernel size & & & \\
    \midrule
    $1 \times 1$     &  98.1 &  \textbf{1456} (357) &  \textbf{94.6} (92.0) \\
    $3 \times 3$     &  \textbf{98.4} &  1540 (\textbf{261}) &  94.4 (\textbf{93.6}) \\
    $5 \times 5$     &  98.3 &  1639 (318) &  94.2 (92.6) \\
    \end{tabular}
\end{table}


\subsection{Influence of the extended MaSTr1478}

Finally, several experiments were performed to evaluate the contribution of the extended training dataset MaSTr1478. In particular, how much performance improvement is brought by the temporal extension and how much by the new scenes with reflections and glitter.  The results in Table~\ref{tab:wasr_comp} show that the single-frame WaSR does not benefit from the additional sequences in MaSTr1478. While the overall detection performance improves by 0.1 points F1, the performance decreases by 0.6 points inside the danger zone. 
Using only temporally extended MaSTr1325 does not improve WaSR-T performance. However, considering also the new sequences in MaSTr1478, the performance improves substantially. We observe a 41\% overall reduction in the number of FP detections and a 53\% reduction of FPs inside the danger zone. The overall performance is thus increased by 1.0 F1 points overall and by 5.4 F1 points inside the danger zone. 

Figure~\ref{fig:cmp_mods} provides qualitative results. In contrast to \mbox{WaSR-T}, the single-frame WaSR is unable to correctly segment regions of water containing the reflections and glitter, despite using the reflection-specific training examples of MaSTr1478. We conclude that both the new scenes and the temporal extension allow learning of the temporal appearance in WaSR-T and are responsible for improved segmentation.


\begin{table}[htb]
    \centering
    \caption{Influence of training dataset extensions in terms of water-edge detection robustness ($\mu_R$), number of FP detections and F1 score. Performance inside the danger-zone is reported in parentheses.}
    \label{tab:wasr_comp}
    \begin{tabular}{lccc}
    model        & $\mu_R$  &    FP &       F1 \\
    \midrule
    WaSR (MaSTr1325)       &  97.2 &  2625 (561) &  93.4 (88.2) \\
    WaSR (MaSTr1478)   &  97.8 &  2492 (629) &  93.5 (87.6) \\
    WaSR-T (MaSTr1325)     &  97.5 &  2273 (655) &  93.7 (87.3) \\
    WaSR-T (MaSTr1478) &  \textbf{98.4} &  \textbf{1540} (\textbf{261}) &  \textbf{94.4} (\textbf{93.6}) \\
    \end{tabular}
\end{table}

\section{Conclusion}

We presented WaSR-T, a novel maritime obstacle detection network that harnesses the temporal context to improve obstacle detection by segmentation on water regions with ambiguous appearance. We also extended the well-known training dataset MaSTr1325~\cite{Bovcon2019Mastr} by including preceding images for each training image and added new 153 training images with challenging scenes containing object mirroring and glitter -- the new dataset is called MaSTr1478. Experiments show that the new images and temporal extension lead to substantial improvement on maritime obstacle detection. WaSR-T outperforms single-frame maritime obstacle detection networks as well as other networks that use temporal contexts and sets a new state-of-the-art on the maritime obstacle detection benchmark MODS~\cite{Bovcon2020MODS}.


\addtolength{\textheight}{-12cm}   






\bibliographystyle{IEEEtran}
\bibliography{references}

\end{document}